\documentclass[10pt,twocolumn,letterpaper]{article}

\usepackage[pagenumbers]{wacv} % To force page numbers, e.g. for an arXiv version

\usepackage{graphicx}
\usepackage{amsmath}
\usepackage{times}
\usepackage{epsfig}
\usepackage{amssymb}
\usepackage{booktabs}
\usepackage{multirow}
\usepackage{paralist,tabularx}
\usepackage{enumitem}
\usepackage[accsupp]{axessibility}  % Improves PDF readability for those with disabilities.
\usepackage{algpseudocode}
\usepackage{algorithm}
\usepackage[pagebackref,breaklinks,colorlinks]{hyperref}

\usepackage[capitalize]{cleveref}
\crefname{section}{Sec.}{Secs.}
\Crefname{section}{Section}{Sections}
\Crefname{table}{Table}{Tables}
\crefname{table}{Tab.}{Tabs.}

\newcommand{\OCRD}{SCoRD\xspace}
\newcommand{\OCRDNet}{SCoRDNet\xspace}

\def\wacvPaperID{2198} % *** Enter the WACV Paper ID here
\def\confName{WACV}
\def\confYear{2024}

\begin{document}

%%%%%%%%% TITLE - PLEASE UPDATE
\title{SCoRD: Subject-Conditional Relation Detection with Text-Augmented Data}

\author{Ziyan Yang$^{1}$\thanks{A portion of this work was done during Ziyan Yang’s internship at Adobe Research.} , Kushal Kafle$^2$, Zhe Lin$^2$, Scott Cohen$^2$, Zhihong Ding$^2$,  Vicente Ordonez$^1$ \\
$^1$Rice University, $^2$Adobe Research\\
\tt\small
\{zy47,vicenteor\}@rice.edu, \{kkafle,zlin,scohen,zhding\}@adobe.com}

% \author{First Author\\
% Institution1\\
% Institution1 address\\
% {\tt\small firstauthor@i1.org}
% % For a paper whose authors are all at the same institution,
% % omit the following lines up until the closing ``}''.
% % Additional authors and addresses can be added with ``\and'',
% % just like the second author.
% % To save space, use either the email address or home page, not both
% \and
% Second Author\\
% Institution2\\
% First line of institution2 address\\
% {\tt\small secondauthor@i2.org}
% }
\maketitle

%%%%%%%%% ABSTRACT
\begin{abstract}
We propose Subject-Conditional Relation Detection (\OCRD), where conditioned on an input subject, the goal is to predict all its relations to other objects in a scene along with their locations.  Based on the Open Images dataset, we propose a challenging OIv6-\OCRD benchmark such that the training and testing splits have a distribution shift in terms of the occurrence statistics of $\langle$subject, relation, object$\rangle$ triplets. To solve this problem, we propose an auto-regressive model that given a subject, it predicts its relations, objects, and object locations by casting this output as a sequence of tokens. First, we show that previous scene-graph prediction methods fail to produce as exhaustive an enumeration of relation-object pairs when conditioned on a subject on this benchmark. Particularly, we obtain a recall@3 of 83.8\% for our relation-object predictions compared to the 49.75\% obtained by a recent scene graph detector. Then, we show improved generalization on both relation-object and object-box predictions by leveraging during training relation-object pairs obtained automatically from textual captions and for which no object-box annotations are available. Particularly, for $\langle$subject, relation, object$\rangle$ triplets for which no object locations are available during training,  we are able to obtain a recall@3 of 33.80\% for relation-object pairs and 26.75\% for their box locations.
\end{abstract}
\vspace{-0.3in}
%%%%%%%%% BODY TEXT
\section{Introduction}
\label{sec:intro}
Scene Graph Generation (SGG) in computer vision commonly refers to the task of predicting the class and locations for all possible objects along with the relations between them. A variety of methods have been introduced in the past years that attempt this task with increasingly accurate results~\cite{johnson2015image,lu2016visual,dai2017detecting, zhuang2017towards, zhang2017visual,yu2017visual,yin2018zoom,zellers2018neural, zhang2019graphical, abdelkarim2021exploring, chen2021reltransformer,cong2022reltr,qi2018learning}. However, predicting or annotating every possible $\langle$subject, relation, object$\rangle$ triplet for every subject in an image, especially under an open-vocabulary setting becomes increasingly impractical. In this work, we instead focus on subject-conditional relation detection (\OCRD) and show that we can generalize to an open-vocabulary setting through extra text annotations. 

%On one end, a variant of this task consists exclusively of predicting the predicates given object labels and locations (i.e.~predicate classification), and at the other end, the task could be about predicting object labels, locations and predicates simultaneously (i.e.~scene graph detection). In all cases, the task consists in considering all objects in a scene, regardless of relative importance, and finding as many pairwise relations between objects, also referred to as predicates. There has been significant efforts in addressing this family of tasks under a variety of settings.  % and output objects locations simultaneously from images. 
%Traditional Scene Graph Generation has been formulated as either Predicate classification -- PredCLS: Given ground truth labels and boxes for subjects and objects, predict the predicates between them; Scene graph classification -- SGCLS: Given ground truth boxes, predict subjects, objects and predicates; or Scene graph detection -- SGDET: Predict boxes and labels for subjects, objects, and predicates from images. 
%However, many previous SGG methods rely on a first stage where pre-trained object detectors are used to generate proposals for objects, thereby limiting relation predictions only among objects in the vocabulary of object detectors. Even though some of the state-of-the-art object detectors such as GLIP~\cite{li2022grounded} and MDETR~\cite{kamath2021mdetr} could localize unseen objects with corresponding text inputs, it is still unclear whether such benefits can be adapted to SGG directly. 

% uncomment here
\begin{figure}[t]
\begin{center}
\includegraphics[width=0.47\textwidth]{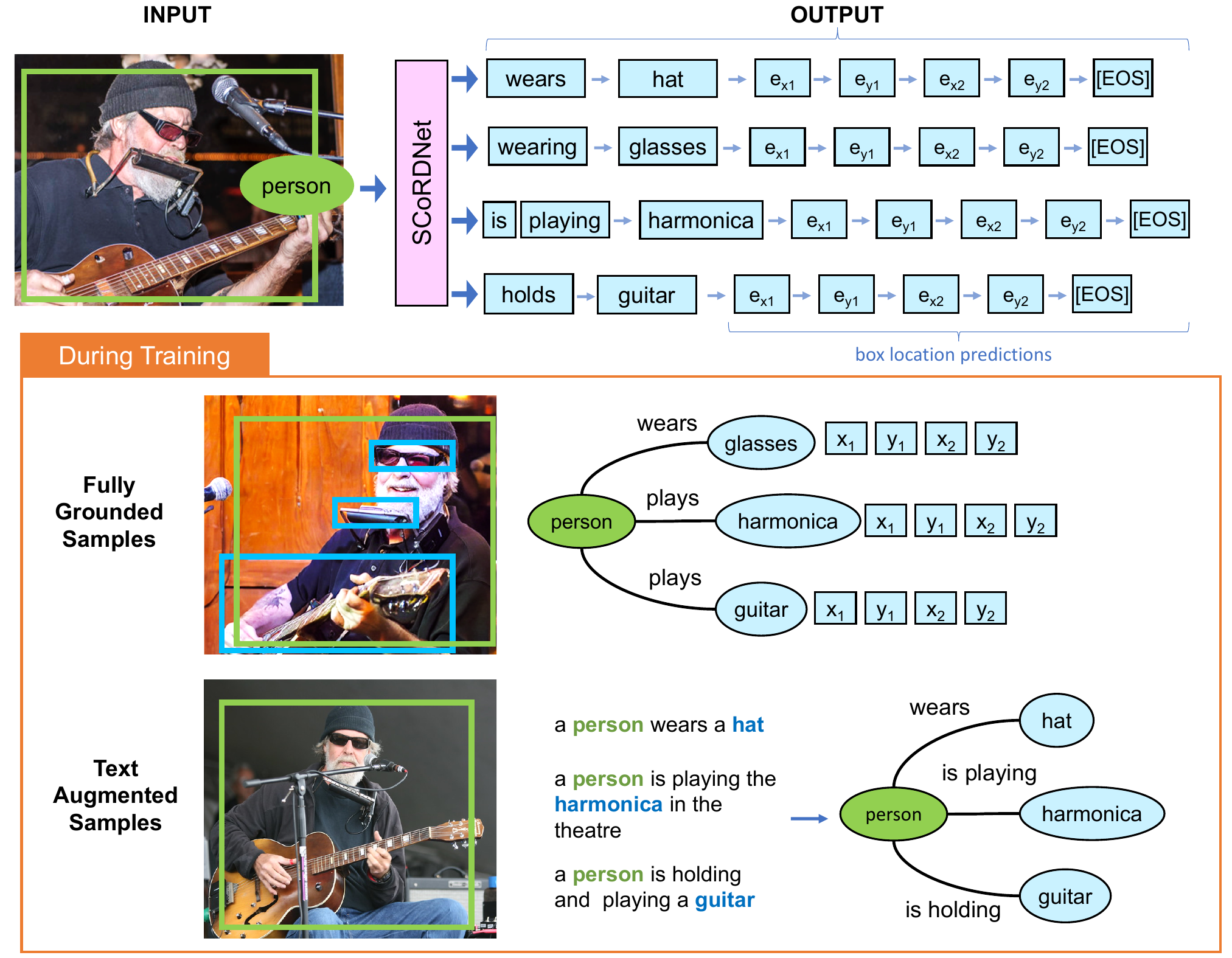}
% \vspace{0.08in}
\vspace{-0.1in}
\caption{We cast subject-conditional relation detection (\OCRD) as a sequence decoding task where given an input subject in an image, we predict relation-object predicates and their locations. We also demonstrate how to leverage ungrounded training samples extracted from parsing textual captions. These samples are easier to obtain than fully grounded samples and have a potentially wider coverage of relation-object pairs. }
\label{fig:lead}
\end{center}
\vspace{-0.35in}
\end{figure}

Many pairwise relations between objects in a scene might be unnecessary for applications where we want to focus only on a given object of interest --a subject-- and its predicates. Consider an image such as the one shown at the top of Figure~\ref{fig:lead}, if we want to crop the person in this picture to create a composite in a different background, we might want to crop the person along with objects that the person interacts with, such as the guitar, the harmonica, the hat, and the glasses. We denote the task of finding objects and predicates that relate to a given subject as Subject-Conditional Relation Detection (\OCRD). We propose a challenging benchmark based on the Open Images dataset such that models trained on our provided training split can not easily take advantage of priors in the highly skewed distribution of $\langle$subject, relation, object$\rangle$ triplets. We provide a strong baseline and additionally demonstrate how we can tackle the prediction of relation-object pairs that are unseen during training by leveraging an external dataset containing text annotations.

\begin{figure*}[h!]
% \vspace{-0.15in}
\begin{center}
\includegraphics[width=\textwidth]{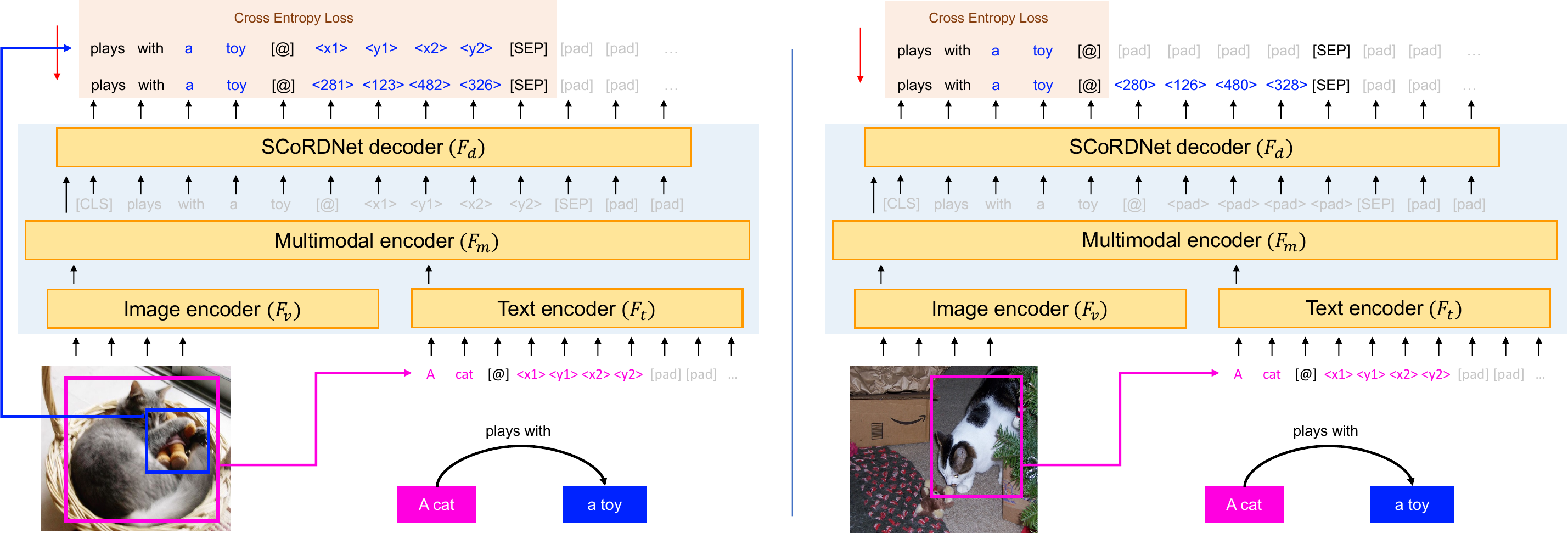}
%\vspace{0.1in}
\caption{Here we show an overview of \OCRDNet. On the left we show how we handle fully grounded training samples consisting of a $\langle$subject, relation, object$\rangle$ triplet for which a box location is available for both the subject and the object. On the right, we show how we handle a training sample for which we only have a $\langle$subject, relation, object$\rangle$ triplet but no object location is available. In the second case, we do not backpropagate gradients through the prediction heads corresponding to object location coordinates although the model often makes reasonable predictions based on parameter sharing with other similar samples that are fully annotated. }
\label{fig:model}
\end{center}
\vspace{-0.26in}
\end{figure*}

We propose \OCRDNet, a model that is trained to predict given an input image and a subject, a relation-object pair along with the object location auto-regressively. At test time, we use a Two-step decoding approach to generate a diverse set of relation-object pairs for a given test input image-subject pair. Figure~\ref{fig:model} shows an overview of this model. \OCRDNet can leverage during training images annotated only with textual captions to improve its capabilites on rare or unseen $\langle$subject, relation, object$\rangle$ triplets. For example, even when our training data contains no annotations for {\em holding umbrella}, we can augment the training data with image-text data to provide an un-grounded example for {\em holding umbrella}; Due to our unified decoding, our model has the ability to not only predict but also provide accurate grounding in the form of a bounding box location for {\em holding umbrella} during inference. Figure~\ref{fig:lead} shows an overview of the inputs and outputs of our model during test time, as well as the two types of annotated samples that it can leverage during training: Fully grounded samples where relation-object pairs as well as object locations are provided, and text-augmented samples where the images are also annotated with image captions, which potentially mention a larger set of object categories than datasets that are fully annotated with $\langle$subject, relation, object$\rangle$ triplets. %Figure~\ref{fig:lead} shows an input to our model (left) and the outputs for our base object-centric relation detection model (middle), and finally the outputs of our model when trained with additional purely textual data. Our model is particularly effective in  discovering relations that were rare or unseen during training but that were mentioned in text corpora. In this example, the base model can detect that the person is feeding a bird, but is not able to detect more exhaustively some of the other relations present such as the object that the person is stepping on.%, where the goal is to predict in an image all the related objects along with their locations conditioned on an input object. 

Our contributions can be summarized as follows:
\begin{itemize}[nolistsep,noitemsep, leftmargin=*]
%\vspace{-0.05in}
\item We propose a new setting for relationship prediction, called Subject-Conditional Relation Detection (\OCRD), which consists of enumerating relations, objects, and boxes conditioned on an input subject in an image.
%\vspace{-0.05in}
\item We propose OIv6-\OCRD, a challenging benchmark that contains a training split and a set of testing splits, such that models trained on this dataset are not able to easily take advantage of the statistical distribution of  $\langle$subject, relation, object$\rangle$ triplets. 
%\vspace{-0.05in}
\item We propose \OCRDNet and conduct extensive experiments under different levels of supervision, demonstrating that our model can ground objects for new relation-object pairs with only additional text supervision. Moreover, we compare to prior general scene-graph generation (SSG) models by conditioning them on a subject and demonstrate the advantages of \OCRDNet.  
\end{itemize}

% Recently, multimodal vision and language learning area has developed rapidly and attracted more attention because of various real-world applications require the joint knowledge coming from different modalities. For example, in image captioning, the model should potentially transfer information from images to text, so the generated descriptions could describe contents in corresponding images. However, many previous methods for object detection and scene graph generation cannot jointly 

\section{Related Work}
%In this section, we will first review related tasks such as Visual Relation Detection (VRD) and Human-Object Interactions (HOI).
% and Human-Centric Relation Segmentation (HRS). 
%Then we will give a survey of the areas of unified sequence prediction and vision-language pretraining. We will discuss the differences and similarities between previous tasks and methods.
Our work is related to prior work on general scene-graph detection, and human-object interaction, and in terms of technical contributions to a few works on general object detection by sequence prediction and vision-language pre-training with grounded inputs.

%\vspace{0.02in}
\noindent{\bf Visual Relation Detection.}
In this task, the model is given an image and expected to produce outputs consisting of all the subject-predicate-object triplets along with their bounding boxes. There have been many works tackling this challenging problem~\cite{lu2016visual,dai2017detecting, zhuang2017towards, zhang2017visual,yu2017visual,yin2018zoom,zellers2018neural, zhang2019graphical, abdelkarim2021exploring, chen2021reltransformer}. Many existing frameworks rely on two steps: First, using object detectors to extract candidate objects with bounding boxes; and then, proposing specific methods to predict relations between objects. 
%Lu et.al.~\cite{lu2016visual} uses RCNN to generate object proposals and designs language and visual learning modules to predict relations; \cite{dai2017detecting} utilizes Faster-RCNN to obtain object proposals and combine regions appearance features and spatial masks to rank possible (subject, relation, object) triplets. 
%The two-step pipeline can be seen as a further step combining basic visual recognition tasks such as object detection and relation recognition. %However, this pipeline requires pre-trained object detectors, therefore, the detected object proposals are limited by the detectors. 
A recent method, RelTR \cite{cong2022reltr} uses a single stage by integrating an end-to-end object detector. %This one-step approach speeds up inference times as well. 
%Our Object-Centric Relation Detection (OCRD) is different from VRD because we are not interested in all the relations and objects in images. Our proposed model will consider a given object and all objects that relate to it. 
Our work focuses instead on selective but exhaustive prediction of relation-object pairs conditioned on a subject.  In order to compare with the previous literature, we use publicly available implementations of Neural Motif~\cite{zellers2018neural} and RelTR~\cite{cong2022reltr} and evaluate them in a subject conditioned setting. We compare against these representative one-stage and two-stage scene graph prediction methods by re-training our model under the same training conditions.  
%\vspace{0.02in}
The Human-Object Interaction Detection task (HOI-det) is a special case of Visual Relation Detection when the subjects are people~\cite{qi2018learning,gkioxari2018detecting,gao2018ican,wang2019deep,hou2020visual,hou2021detecting,hou2021affordance,hou2022discovering, fang2021dirv,wang2022learning}. %This task requires the detection of more fine-grained information between objects instead of more general relations as in VRD. Similar to methods for VRD, previous methods for HOI-det also rely on object detectors to predict box proposals for subjects and objects. 
%For example, \cite{gkioxari2018detecting} uses Faster-RCNN to construct object detection branch, and combines with human-centric branch and interaction branch to predict final results. \cite{wang2019deep} employs FPN to detect human and object proposals. 
%Inspired by CLIP, the recent method of Wang~et~al~\cite{wang2022learning} formulates HOI-det as image-to-text matching and proposes a transformer-based detector to solve the task, showing the ability to handle unseen interactions. Attention-based methods are explored to solve this task as well. For example, Gao~et~al~\cite{gao2018ican} designs an instance-centric attention module to highlight regions that can improve HOI-det. HOI-det has also been tackled using Graph Neural Networks (GNN) as structural representations~\cite{qi2018learning}. %infers a parse graph and labels to represent HOI-det results. 
Our \OCRD task also aims to detect related objects given an input subject but our subjects are not limited to instances of people. 
% \subsection{Human-Centric Relation Segmentation}
% The difference between HOI-det and Human-Centric Relation Segmentation (HRS) is the latter one aims to generate pixel-level segmentation masks for humans and objects instead of bounding boxes, and can be considered as a more fine-grained case of HOI-det. This new task is proposed by Liu et.al. \cite{liu2021human}, and they also designs a model composed of a feature extractor and three branches to segment humans, objects and predict relations. Our proposed task is different from this task because our task supposes the model knows the subjects and the subject locations. 

%\vspace{0.02in}
\noindent{\bf Object Locations as Tokens.}
%To solve generation-based NLP tasks such as image captioning, text generation and machine translation, models usually are designed with an encoder-decoder architecture~\cite{chen2022obj2seq, chen2021pix2seq,bahdanau2015neural,tan2019text2scene}. 
%The decoder is used to generate a sequence of tokens as outputs.% For image generation, the sequence prediction generates outputs pixel by pixel or patch by patch. Previous works such as  and \cite{chen2022obj2seq} attempt to reformulate computer vision tasks as generation-based tasks.
Our proposed \OCRDNet model leverages auto-regressive sequence prediction and predicts relation-object pairs and box locations by considering the box locations as additional discrete output tokens in a sequence. Some recent works have shown the effectiveness of encoding object box coordinates as discrete tokens~\cite{tan2019text2scene, chen2022obj2seq, yang2022unitab,yao2022pevl}. Tan~et~al~\cite{tan2019text2scene} predicts box locations to compositionally generate a scene by discretizing the output space. Chen~et~al~\cite{chen2022obj2seq} considers object detection as a language modeling task by generating bounding box coordinates as text tokens conditioned on input images and previously predicted tokens. %By providing different prompts, this approach can generate corresponding task-specific sequences. 
UniTAB~\cite{yang2022unitab} solves multiple tasks by using both text and box tokens as supervision. 
%For example, by providing UniTAB with ``A picture of'' and one image, it can generate objects classes and bounding box tokens for the generated objects. These branch of works show the possibility of unifying bounding box coordinates and text tokens into the same format, but for our proposed task, the model needs to unify both inputs and outputs from different modalities. 
In the context, of vision-language pre-training, Yao~et~al~\cite{yao2022pevl} recently proposed PEVL which  incorporates grounded input text in its training image-text pairs by encoding object box locations as discrete tokens. 
We build upon this idea to incorporate a decoding transformer that also predicts discrete object-location tokens.

\noindent{\bf Visual Grounding.} These methods aim to localize an arbitrary input text given an input image and are often framed as referring expression comprehension~\cite{kazemzadeh2014referitgame,plummer2015flickr30k,nagaraja2016modeling,yu2018mattnet,qiao2020referring,gupta2020contrastive,arbelle2021detector}. Recent methods such as GLIP~\cite{li2022grounded} and MDETR~\cite{kamath2021mdetr} have impressive general capabilities in their ability to localize arbitrary textual input and can be used as open vocabulary object detectors. Our work also aims to localize objects, however, our goal is to selectively detect objects that are related to a given subject, and also assign a label to the relation. In one of our experiments, we adopt GLIP to produce noisy annotations for input subjects in images that only have textual caption annotations and for comparing our method against a recent work that predicts relations but does not localize objects~\cite{pham2022improving}.

\section{Problem Definition}

%The goal of the \OCRD task is to enumerate as many $\langle$relation, object$\rangle$ pairs for a given subject, along with the box locations corresponding for each object. The outputs of models for this task should include relations and objects such as ``holding umbrella'', which we refer to as relation-object pairs, in addition to the bounding box for predicted objects. %The input is an object query, provided in the form of the bounding box and the name of the subject of interest.

Let the inputs to our task be image $I$ and target subject $s$ along with its corresponding box location $b_{s}$. Our goal in this work is to predict a set of all possible relations, object, and object locations: 
\begin{equation}
\{\langle r, o, b_{o}\rangle \textit{ s.t. }\exists \langle s, r, o\rangle \in \Pi_I \},
\end{equation}

where $\Pi_I$ is an enumeration of all valid relation triplets in input image $I$, and $b_{o}$ is the box location corresponding to object $o$. In this work, we propose a model $\Phi$ such that we can sample output relation-object predictions from a model trained to predict a distribution over all possible values as follows:
\begin{equation}
    \langle r, o, b_{o}\rangle \sim \Phi(I, s, b_s).
\end{equation}

By sampling multiple times from this model, we can obtain an arbitrary set of predictions $\{\langle r_j, o_j, b_{o_j}\rangle\}$. For reference, scene graph generation methods aim to generate a full list of subject, relation, objects along with their locations, i.e.~$\langle s, b_s, r, o, b_{o}\rangle$. However, the general scene graph generation (SGG) setup often limits the number of object categories for $s$, $r$, and $o$. In our setup, $s$, $r$ and $o$ are open-vocabulary and modeled as text, so they can potentially support an arbitrary set of objects and relations.

%The model should predict relation-object pairs $\{r_j, o_j\}_{j=1}^M$, 
%objects $\{o_j\}_{j=1}^M$ 
%and object locations $\{b_{o_j}\}_{j=1}^M$ for each corresponding subject $s_i$ and its location $b_{s_i}$. 

%Our proposed task is different from the scene graph generation (SGG) task. Previous methods for SGG usually do not consider subject locations as inputs, and try to predict all objects and possible relations between objects to rank $\langle$subject, relation, object$\rangle$ triplets based on prediction scores. The general approach for SGG typically requires a pre-trained object detector to generate proposals for objects and often does not ground objects directly. Such an approach is limited by pre-defined object and relation classes and is difficult to leverage training data with different levels of supervision, e.g., data without bounding boxes for objects. In our task, we cast the relation-object prediction and grounding problem as a text generation problem and demonstrate its ability to combine both grounded and un-grounded object instances for training.

\section{Benchmark}
\label{sec:benchmark}
We construct a test benchmark for subject-conditional relation detection based on the Open Images v6 (OIv6) dataset~\cite{kuznetsova2020open} which contains fully grounded $\langle$subject, relation, object$\rangle$ triplets for a diverse array of object categories. We select training and testing splits so that models trained on this benchmark can not rely on the highly skewed statistics of naturally occurring $\langle$subject, relation, object$\rangle$ triplets. 

\subsection{Data}
We use two types of supervision: First, fully-grounded data that contains $\langle$subject, relation, object$\rangle$ triplets with corresponding box locations for both subjects and objects, and a second type of samples that contain noisy $\langle$subject, relation, object$\rangle$ triplets extracted automatically from textual image captions but for which no object locations are available (ungrounded samples). %Next, we describe in detail each of these:

\vspace{0.05in}
\noindent\textbf{Fully-grounded Data:}
We use a combination of Visual Genome~\cite{krishna2017visual}, Flickr30k~\cite{plummer2015flickr30k} and Open Images V6 (OIv6)~\cite{kuznetsova2020open}. Visual Genome has 100k images annotated with 2.3M relations between objects with annotated box locations. Flickr30k contains 30k images with bounding boxes annotated with corresponding referring expressions. We use a language dependency parser~\cite{honnibal2020spacy} to extract $\langle$subject, relation, object$\rangle$ triplets from these expressions, and map them to phrase bounding boxes. Open Images has 526k images with annotations for subjects, objects, relations and boxes. However, many of the relation-object pairs are  relations of the type ``A is B'' which is not a true relation between two different objects. We filter out images with such relations and end up with 120k images. In the following sections, {\em fully grounded data (FG)} is used to refer to any sample from the three sets described in this section.

\vspace{0.05in}
\noindent\textbf{Text-Augmented Data:}
While {\em fully grounded data} is scarce and hard to annotate, it is considerably easier to obtain images annotated with textual captions. We aim to show how automatically extracted $\langle$subject, relation, object$\rangle$ triplets from captions can help on this task even when no box annotations are provided.
%Our training allows easy expansion of image-caption data as an additional signal. 
We use two  data sources: COCO~\cite{lin2014microsoft} and Conceptual Captions (CC3M~\cite{sharma2018conceptual} + CC12M~\cite{changpinyo2021conceptual} referred together as CC). COCO includes 120k images, 5 captions for each image, and bounding box annotations for 80 categories. Despite having box annotations, the COCO dataset does not contain a mapping between object annotations and references to objects in text descriptions. Moreover, object boxes are only annotated for 80 object categories, which does not cover the full range of objects mentioned in captions. Hence, we use COCO as an {\em ungrounded} data source. Additionally, we augment our use of COCO with Localized Narrative descriptions~\cite{PontTuset_eccv2020} to get more varied subjects, objects and relations.
% For example, COCO has ``person'' and bounding boxes for ``person'', but in captions, annotators describe images using words such as ``woman, man, girl, boy'' to indicate ``person'' instead of using ``person'' directly. Second, there can be multiple instances of the same object. And finally, only the 80 categories have boxes, which severely limits the vocabulary of objects. Furthermore, we are not interested in whether additional fully-grounded data can help, but in whether additional image-text data alone can help improve relation prediction and grounding. 
This combination of ungrounded image-text data from COCO and CC is meant to provide additional supervision from an external source. Using image-text pairs, we can extract $\langle$subject, relation, object$\rangle$ triplets with a dependency parser~\cite{honnibal2020spacy}. We also generate noisy box locations for the input subjects in these triplets using GLIP~\cite{li2022grounded} but leave target objects ungrounded without any corresponding box locations.

\subsection{Test-Train Splits}
%We are interested in three main questions about text augmentation. Can additional image-text data: (1) help improve overall performance?
 %(2) help improve performance on relation-object pairs that do not have enough samples during training? and (3) allow for prediction of relation-object pairs that do not have \textit{any} fully grounded training samples? However, the training and validation sets for Open Images V6 follow the same distribution. 
Our test splits are all based on the Open Images v6 validation and testing sets which contain 7,000 images with 20,000 $\langle$subject, relation, object$\rangle$ triplets. Evaluating visual relation prediction tasks is challenging as some relations such as $\langle$sitting on, chair$\rangle$ are overly represented in the dataset to an extent that a model could be relying entirely on dataset statistics. To avoid models from taking advantage of the highly skewed distribution of $\langle$subject, relation, object$\rangle$ triplets and in order to test whether models can generalize to under-represented $\langle$subject, relation, object$\rangle$ triplets, we devise two test splits designed to under-sample the triplets included in the training splits. 
 %Without collecting additional annotations, it is difficult to answer all of these questions. 
%We devise the following scheme:
First, we select all relation-object pairs from the parsed captions obtained from COCO and CC that occur between 100 and 20,000 times. Among those, 104 relation-object pairs also appear in the test split of Open Images V6. We divide these unique 104 unique relation-object pairs into two non-overlapping sets which we denote as {\em Rel-Obj Set A}, and {\em Rel-Obj Set B}, each with 52 unique relation-object pairs. Table~\ref{table:dataset} shows a detailed count of how many samples exist for each of these relation-object pairs across all our data sources. Next, we describe our testing and training splits:

%For these 104 relation-object pairs, we manipulate the training dataset to simulate the following two scenarios:

%\vspace{0.05in}
%\noindent
%\textbf{Full Test: }
%We evaluate our methods on the Open Images V6 validation and testing sets which contain annotations for subject labels, object labels, relation labels, and bounding boxes. There are 7,000 images in these sets with 20,000 subject-relation-object triplet samples in total. 

%Our model takes a subject name and the bounding box for this subject as inputs and generates relations, objects, and bounding boxes for objects as outputs. 

\vspace{0.05in}
\noindent
\textbf{Test A: } %We select 52 relation-object pairs from the aforementioned 104 pairs, and partially remove training samples from the fully-grounded training triplets. %In Figure~\ref{fig:main}, ``riding boat'' is such a relation-object pair.
%This scenario simulates not having enough grounded data during training. Performance on these relation-object pairs aims to test whether adding text-only data can reinforce previously seen but under-represented relation-object pairs.
This test split consists of $\sim$4k samples belonging to any of the $\langle$subject, relation, object$\rangle$ triplets covered by the 52 unique relation-object pairs in the {\em Rel-Obj Set A} set. %However, for the training split we . 

\vspace{0.05in}
\noindent
\textbf{Test B: } This test split includes $\sim$4k samples belonging to any of the $\langle$subject, relation, object$\rangle$ triplets covered by the 52 unique relation-object pairs in the {\em Rel-Obj Set B} set. 

%We select the remaining 52 relation-object pairs from the 104 pairs, and remove training samples from the fully-grounded training triplets belonging to these pairs. 
%In Figure~\ref{fig:main}, ``holding paddle'' is such a relation-object pair. 
%For example, we remove all examples of the triplet  $\langle$[subject], on, cake$\rangle$ and corresponding bounding boxes from the training set. This scenario represents not having any grounded data in the training split. Performance on these relation-object pairs will demonstrate whether adding text-only data can help compositional generalization.% and transfer knowledge about grounding to predict and ground relation-object pairs that have no fully-grounded subject-relation-object triplet data.

\vspace{0.05in}
\noindent
\textbf{Full Test:} This test split includes the full test set samples of OIv6 containing $\sim$20k fully grounded samples $\langle$subject, relation, object$\rangle$ triplets.  

\vspace{0.05in}
\noindent
\textbf{Base Training Split:} Consists of fully grounded training samples. However, we remove half of the samples belonging to relation-object pairs present in {\em  Rel-Obj Set A}: $\sim$11.5k from VG, $\sim$2.5k from Flickr30k and $\sim$34k from OIv6 (Table~\ref{table:dataset}). Our goal is to skew the distribution between the training and test splits so that the relation-object pairs present in the test split of {\em Test A} are underrepresented in the training split. We additionally remove all the samples that include triplets containing relation-object pairs present in {\em Rel-Obj Set B}: $\sim$37k from VG, $\sim$5k from Flickr30k and $\sim$104k from OIv6, (Table~\ref{table:dataset}). This will make {\em Test B} challenging since the relation-object pairs in this part of the test set will not be present in this training split.

\vspace{0.05in}
\noindent
\textbf{Text-Augmented Training Split:} This training split includes all the fully grounded samples in the base training split and additionally includes ungrounded samples that were automatically annotated from textual captions: $\sim$85k from COCO and $\sim$94k from CC. These samples were obtained by filtering COCO and CC to target the 104 relation-object pairs targeted by both {\em Test A} and {\em Test B}. The goal is to demonstrate compositional generalization from ungrounded samples to grounded but underrepresented (Test A), and unseen $\langle$subject, relation, object$\rangle$ triplets (Test B). 
% We plan to release training and test splits across all these scenarios as well as code upon paper acceptance.

% We plan to release training and test splits across all these scenarios as well as code to compute evaluation metrics across our three test splits under all the described scenarios with both fully grounded training data and fully grounded + text augmented data. 

\begin{table}[t!]
\centering
\setlength\tabcolsep{1pt}
\renewcommand{\arraystretch}{1.1}
 \begin{tabular}{l c c c c c c c} 
 \toprule
 %\textbf{split} 
 %&&
 \textbf{Source} & \textbf{Obj-Loc} &&
 \begin{tabular}{@{}c@{}}
 \textbf{Rel-Obj} \\ 
 \textbf{Set A}\end{tabular} 
 &&
  \begin{tabular}{@{}c@{}}
 \begin{tabular}{@{}c@{}}
 \textbf{Rel-Obj} \\ 
 \textbf{Set B}\end{tabular} 
  \end{tabular} 
 && 
 \begin{tabular}{@{}c@{}}\textbf{Full} \\ \textbf{Rel-Obj Set}\end{tabular}  
 \\ [0.5ex] 
 \midrule
 
%\multirow{5}{*}{Train}&& 
VG~\cite{krishna2017visual} 
& \checkmark && 23k && 37k && 2M \\ 

 %&&
 Flickr30k~\cite{plummer2015flickr30k} & 
 \checkmark && 5k && 5k && 277k \\
 
 %&&
 OIv6~\cite{kuznetsova2020open} 
 & \checkmark && 68k && 104k && 350k \\
 %\midrule
 
 %&&
 COCO~\cite{lin2014microsoft,chen2015microsoft}* & - && 49k && 36k && 85k \\
 %&&
 CC~\cite{sharma2018conceptual,changpinyo2021conceptual}* & - && 38k && 56k && 94k \\
 \midrule
 
%Test && 
OIv6 (test)~\cite{kuznetsova2020open} & \checkmark && 4k && 4k && 20k \\

\bottomrule
 \end{tabular}
 \vspace{-0.08in}
 \caption{\textbf{Rel-Obj Set A} focuses on a set of 52 unique relation-object pairs, and \textbf{Rel-Obj Set B} focuses on a non-overlapping set of 52 unique pairs. This table shows how many individual samples exist in each of the data sources we use to construct our training and test splits for each of these relations. (*) For ungrounded text-augmented data sources we only consider samples included in the 104 unique relation-object pairs in Test A and B.}
 \vspace{-0.1in}
 \label{table:dataset}
\end{table}

%\vspace{-0.1in}
\section{Method}
\label{sec:method}
In this section, we describe our proposed \OCRDNet model. Figure~\ref{fig:model} shows an overview of our model, as used in two different training modes of supervision. Our model consists of four transformer models, an image encoder, an input text encoder, a multimodal fusion encoder, and an output auto-regressive decoder. Here we describe these components along with our proposed decoding process. % task as well as the data sources used for training our models, including the text-augmentation step to improve the performance of our model. % We outline our model structure, loss functions, and a two-step decoding process that allows us to train on data that includes grounded text, and ungrounded text.
%$$\textbf{Backbone: } 
%Fully supervised relation detection requires subjects, objects, relations, bounding boxes for subjects, and bounding boxes for objects, which can be time-consuming and expensive to annotate. Our model can effectively learn from partially annotated relation-object pairs and an additional un-grounded (i.e. no box annotations) expansion dataset obtained by parsing captions. The two sources of data used for our model are as follows:

\subsection{Model}
\label{sec:model}
We cast the problem of subject-conditional relation detection as a sequence-to-sequence model where the input is an image $I$ coupled with an input token sequence representing the input subject $s$ and its box location $b_s$, and the output is a sequence of tokens representing the predicted relation-object pair $\langle r, o \rangle$ and the corresponding object location coordinates $b_o$. 
The left part of Figure~\ref{fig:model} shows an overview of this process for a given input fully grounded training sample. In order to cast the input and output coordinates for the grounded samples as tokens, it is necessary to discretize them. For a given set of box coordinates ($x_1$, $y_1$, $x_2$, $y_2$) and a corresponding image with height $h$ and width $w$, the goal is to discretize the coordinates as: 
\begin{equation}
(P\frac{x_1}{w}, P\frac{y_1}{h}, P\frac{x_2}{w}, P\frac{y_2}{h}),
\end{equation}
with a pre-defined number $P$ representing the total number of position tokens. Additionally, we add special separator tokens to indicate the start and the end of box coordinates.

\OCRDNet consists of an image transformer encoder $F_{v}$ which encodes images into a sequence of image features $\{I_i\}_{i=1}^{N_I}$, and a text transformer encoder $F_{t}$ which encodes subjects with position tokens as a sequence of text features $\{T_j\}_{j=1}^{N_T}$. Then, a multimodal fusion encoder $F_{m}$ will fuse image and text features through cross-attention layers and produce a context vector $z$.  Finally, the output context vector $z$ of the multimodal fusion encoder $F_{m}$ is forwarded to a decoder transformer $F_{d}$ which is trained auto-regressively to predict a relation, an object, and its corresponding box coordinate locations as a sequence of tokens. 

Our model is trained to predict a context vector \mbox{$z = F_m(F_v(I), F_t(s, b_s))$}, which is then used to predict an output relation-object pair along with a bounding box using the auto-regressive transformer decoder \mbox{$\langle r, o\rangle, b_o = F_d(z)$}. During training, the model is optimized to minimize a loss function $\mathcal{L}(\langle r, o\rangle, b_o, F_d(z))$ that aims to produce a sequence from the transformer decoder that matches the true relation-object pair. During inference, predictions are obtained by sampling from the transformer decoder: $\langle r, o\rangle, b_o \sim F_d(z)$. This same model can be trained with ungrounded relation-object pairs for which a set of object box coordinates is not available by simply optimizing $\mathcal{L}(\langle r, o\rangle, F_d(z))$ for samples that do not contain a box corresponding to the object in the relation. However, our model can still be used to sample a full length sequence containing a relation-object pair as well as the object location. We show on the right part of Figure~\ref{fig:model} an overview of how this process works. 
%it is not necessary to have all types of input tokens available during training. 
%This allows us to use text-only augmentation to learn from ungrounded data sources. 
During training, we can simply skip computing the loss terms for input tokens that are not provided in the ground truth output sequences (e.g., boxes). %This allows us to use ungrounded text-augmentation as described in the earlier section.

\vspace{0.02in}
\noindent
\textbf{Loss: } We use a standard cross-entropy loss to train our model. Let $\mathrm{H}(\cdot,\cdot)$ be the cross-entropy function, then we compute the cross-entropy between the current target token and generated token conditioned on the input image, subject sequence, and previous tokens as follows:
\vspace{-0.02in}
\begin{equation}
    z_i = F_m\left(F_v(I_i), F_t(\langle s_i, b_{s_i}\rangle)\right)
\end{equation}
\begin{equation}
\mathcal{L} = \sum_{i=1}^N\sum_{k} \mathrm{H}\left(t_k, F_d \left(t_{0:k-1}, z_i \right)\right),
\end{equation}
where $N$ is the number of training samples, $I_i$ is the input image, $\langle s_i, b_{s_i}\rangle$ is an input subject and subject-box encoded as a sequence of tokens, and $t_k$ is a token at a given time step $k$ for ground truth annotation $\langle\langle r_i, o_i\rangle, b_{o_i}\rangle$ encoded as a sequence of tokens. For ungrounded samples these tokens $t_k$ are encoding only the ground truth annotation $\langle r_i, o_i\rangle$.

\subsection{Two-step Decoding} 

%\renewcommand{\algorithmicrequire}{\textbf{Input:}}
%\renewcommand{\algorithmicensure}{\textbf{Output:}}
%\begin{algorithm}[t!]
%  \caption{Two-stage decoding}
%  \begin{algorithmic}[1]
%   \end{algorithmic}
 % \label{alg:layer-wise-feedback}
%\end{algorithm}

\renewcommand{\algorithmicrequire}{\textbf{Input:}}
\renewcommand{\algorithmicensure}{\textbf{Output:}}
\begin{algorithm}[t!]
\caption{Two-stage decoding}\label{alg:twostage}
\begin{algorithmic}[1]
\small 
%\Variables
 \Require $I$: Input image
 \Require $\langle s, b_s\rangle$: Input subject, and subject box
 \Require $K$: Number of desired output relation-object pairs
 \Ensure $\{\langle r_k, o_k\rangle\}, b_{o_k}$: Output relation-object pairs and boxes
 
%\EndVariables
%\Statex
\vspace{0.05in}
\Function{BeamSearch}{$F, z, \{t_k\}, K, \text{[EOS]}$}
    \State return top K sequences from $F(z, \{t_k\})$ ending in \text{[EOS]}
\EndFunction
\vspace{0.05in}
%\Statex

\State $z \gets F_m(F_v(I),F_t(\langle s, b_s\rangle))$

\State  $\{\langle r_k, o_k\rangle\} \gets $ \Call{BeamSearch}{$F_d, z, \emptyset, K, [@]$}

\For{$\langle r_k, o_k\rangle$ in $\{\langle r_k, o_k\rangle\}$}
%\State $z \gets z + \langle r_k, o_k\rangle$
\State  $b_{o_k} \gets $ \Call{BeamSearch}{$F_d, z, \langle r_k, o_k\rangle, 1, \text{[SEP]}$}
\EndFor
\end{algorithmic}
% \vspace{-0.08in}
\end{algorithm}
\vspace{-0.08in}
We could sample output relation-object pairs along with object-locations using beam search directly from our decoder such that $\{\langle r, o\rangle\}, b_{o} \sim F_d$. However, we find that this leads to a lack of diversity in the predicted relation-object pairs due to the disparity in the vocabulary for relation-object tokens and box location tokens. Hence, we rely on a Two-step decoding process where first a diverse set of $K$ relation object pairs $\{\langle r_k, o_k\rangle\}$ are decoded, and then a corresponding set of object boxes $b_{o_k}$ are decoded conditional on the relation-object pairs. This two-step decoding process is described in Algorithm~\ref{alg:twostage}, where we rely on a beam search function that decodes a sequence of $K$ tokens given an input decoder $F$, conditional state vector $z$, and a partially generated output sequence ${t_k}$. Additionally, this function takes as a parameter a custom end-of-sequence token. First, we perform beam search to decode relation-object pairs until we encounter the end of sequence token $[@]$ which we use to separate relation-object pairs and object-location box coordinates, and then we decode box locations one by one conditioned on the same input but with a partially generated output sequence comprised of the relation-object pair. We decode until finding the end-of-sequence token [SEP].

\begin{table}[t!]
\centering
\renewcommand{\arraystretch}{1.2}
\begin{tabular}{llrrrrrr}
\toprule
\textbf{} &\textbf{} & \multicolumn{2}{c}{\textbf{Rel-Object}} & \multicolumn{2}{c}{\textbf{Object-Loc}} \\
\cmidrule(l){3-4} \cmidrule(lr){5-6}
\textbf{Split} &\textbf{Model} & {\em R@1} & {\em R@3} & {\em R@1} & {\em R@3} \\
\midrule
\multirow{2}{*}{\textbf{Test A}} & Base & 39.94 & 82.19 & 29.08 & 57.06 \\
 & Text-aug. & 43.61 & 83.91 & 31.75 & 58.58 \\
 \midrule 
\multirow{2}{*}{\textbf{Test B}} & Base & 0.74 & 4.78 & 0.67 & 4.36 \\
 & Text-aug. & 18.71 & 33.80 & 15.14 & 26.75 \\
 \midrule
\multirow{2}{*}{\textbf{Full Test}} & Base & 49.84 & 69.22 & 36.54 & 50.37 \\
 & Text-aug. & 53.48 & 75.51 & 39.36 & 55.38 \\
\bottomrule
\end{tabular}
\vspace{-0.08in}
\caption{Results for our three test splits in our benchmark. The base model was trained while removing 50\% of training samples with triplets in {\em Test A} and by removing all samples for triplets present in {\em Test B}. These results highlight how our Text-augmented model is able to take advantage of ungrounded samples to generalize to both unseen and under-sampled relation-object pairs, not only for predicting their correct relation but also object locations.}
\label{table:maintable}
\vspace{-0.2in}
\end{table}

\section{Experiment Settings}
\label{sec:exp}

% In this section we describe implementation details and evaluation metrics.

\vspace{0.05in}
\noindent\textbf{Implementation Details.}
For the image encoder $F_v$, the text encoder $F_t$ and the multimodal fusion encoder $F_m$ we adopt the architecture and pre-trained encoders of PEVL~\cite{yao2022pevl}, which is a joint vision-language transformer which is in turn based on the ALBEF model~\cite{li2021align} but using additional grounded supervision with box coordinates as input tokens. Our transformer-based model contains one $12$-layer vision transformer\cite{dosovitskiy2020image} as the image encoder, one $6$-layer text transformer~\cite{vaswani2017attention,devlin2018bert} as the text encoder, one $6$-layer multimodal transformer encoder to combine image and text information and one $6$-layer transformer decoder to generate target sequences. The hidden size for each layer is $768$ and the number of attention heads is $12$. The decoder follows the same architecture as the multimodal transformer encoder but using masked self-attention layers, and its parameters are initialized using the weights from the multimodal transformer encoder. Code and data are available\footnote{ \url{https://github.com/uvavision/SCoRD} }.

% OIV6 is a good candidate

\vspace{0.05in}
\noindent\textbf{Evaluation metrics.} 
We report Recall@K for all the experiments. For each subject and its bounding box, we keep different numbers of sequences from beam search -- which is our mechanism for decoding. For example, Recall@3 indicates we keep 3 relation, object, object-boxes with the highest scores. We split the outputs into two parts: relation-object (Rel-Object) and object-location (Object-Loc). 
%For text, we evaluate on relation+object; for the bounding box part, we evaluate on predicted object boxes. 
If one predicted text part among K returned sequences belongs to the ground truth text part or its synsets, we count this sample as positive for text evaluation. If one predicted text is ``correct'', we keep looking at the predicted box, and if the predicted box and ground truth box have IoU $\ge$ 0.5, this sample is counted as positive for the bounding box part.

 \begin{table}[t!]
 \renewcommand{\arraystretch}{1.2}
 \small
 \centering
 \begin{tabular}{l l c c c c} 
\toprule
  
 & & \multicolumn{2}{c}{\textbf{Rel-Object}} & \multicolumn{2}{c}{\textbf{Object-Loc}} \\ \cmidrule(l){3-4} \cmidrule(lr){5-6}
\textbf{IoU}& \textbf{Methods} & {\em R@1} & {\em R@3} & {\em R@1} & {\em R@3} \\ 
  \midrule
  
 \multirow{4}{*}{0.5} & RelTR~\cite{cong2022reltr} & 29.71 & 49.75 & 24.64 & 42.92 \\
 &\OCRDNet  & 65.60 & 83.80 & 24.88 & 40.81 \\
 \cmidrule(l){2-6} 
%  Motif & 83.55 & 63.62 \\
%  Motif & 68.33 & 56.85 \\
%  Ours & 93.65 & 71.54 \\ % our all-grounded model on 33 classes, does not remove anything
 &Motif~\cite{zellers2018neural} & 25.83 & 41.79 & 23.22 & 37.05 \\
 &\OCRDNet & 65.17 & 82.31 & 23.38 & 39.97 \\
 \midrule
 \multirow{3}{*}{0.4} & RelTR~\cite{cong2022reltr} & 29.71 & 49.75 & 26.64 & 44.84 \\
 & \OCRDNet  & 65.60 & 83.80 & 30.71 & 49.02 \\  \cmidrule(l){2-6} 

 & Motif~\cite{zellers2018neural} & 25.83 & 41.79 & 24.17 & 38.55 \\
 & \OCRDNet  & 65.17 & 82.31 & 30.33 & 49.37 \\
 \midrule
 \multirow{4}{*}{0.3} & RelTR~\cite{cong2022reltr} & 29.71 & 49.75 & 27.18 & 46.03 \\
 & \OCRDNet & 65.60 & 83.80 & 35.47 & 55.70 \\ \cmidrule(l){2-6} 

 & Motif~\cite{zellers2018neural} & 25.83 & 41.79 & 29.72 & 39.73  \\
 & \OCRDNet  & 65.17 & 82.31 & 34.60 & 55.61 \\
%  \hline
% %  TAP+GLIP & 63.55 & xxxx  \\ 
%  TAP+GLIP & 16.31/29.18/38.13 & xxxx  \\ 
% %  Ours+GLIP & - & xxxx  \\ 
%  Ours & 63.16/91.53/95.12 & 49.50/71.46/74.34 \\
%  \hline
%  RelTR-0.5 & 29.71/49.75/56.16 & 24.64/42.92/47.68 \\
%  Ours-0.5 & 65.60/83.80/88.83 & 24.88/37.35/38.89 \\
% %  Motif & 83.55 & 63.62 \\
% %  Motif & 68.33 & 56.85 \\
% %  Ours & 93.65 & 71.54 \\ % our all-grounded model on 33 classes, does not remove anything
%  Motif-0.5 & 25.83/41.79/52.76 & 23.22/37.05/46.76 \\
%  Ours-0.5 & 65.17/82.31/86.49 & 25.91/34.99/36.65 \\
%  \hline
%  RelTR-0.4 & 29.71/49.75/56.16 & 26.64/44.84/51.06 \\
%  Ours-0.4 & 65.60/83.80/88.83 & 30.71/49.02/46.99 \\
%  Motif-0.4 & 25.83/41.79/52.76 & 24.17/38.55/48.66 \\
%  Ours-0.4 & 65.17/82.31/86.49 & 31.67/41.55/43.84 \\
%  \hline
%  RelTR-0.3 & 29.71/49.75/56.16 & 27.18/46.03/52.48 \\
%  Ours-0.3 & 65.60/83.80/88.83 & 39.04/51.67/53.70 \\
%  Motif-0.3 & 25.83/41.79/52.76 & 29.72/39.73/49.92  \\
%  Ours-0.3 & 65.17/82.31/86.49 & 37.12/48.02/50.63 \\
 
% only when subject is correct, choose top-k then 

%  \hline 
\bottomrule
\end{tabular}
\vspace{-0.08in}
\caption{Comparison with general scene-graph generation models on OIv6-\OCRD. 
%In order to compare fairly to these methods, we calculate subject-conditional results for previous SSG models by only considering test samples for which the corresponding SGG models correctly predict and localize the subjects, and compute our conditional results for the same test samples. 
\OCRDNet compares favorably against the strong Motif model~\cite{zellers2018neural} and the recently proposed RelTR~\cite{cong2022reltr}. For this experiment all methods were trained on Visual Genome~\cite{krishna2017visual}.}
\label{table:compare_sgg}
\vspace{-0.2in}
\end{table}

% \subsection{Training Details}
\vspace{-0.1in}
\section{Results and Discussion}
\label{sec:results}
% \textbf{enhancement set: } This subset includes 52 high-frequency relation-object pairs existing in both OIV6 validation/testing sets and OOD datasets. We filter out different portions of relation-object (0\%, 25\%, 50\%, 75\%) with boxes samples for these 104 relation-object pairs from in-domain data. Then, we add OOD data samples consist of subjects with bounding boxes and relation-object pairs for the enhancement set. We conduct experiments on the enhancement set to demonstrate the possibility to obtain improved relation detection results on relation-object pairs that have partially grounded data. 

% \textbf{expansion set: } This subset includes another 52 high-frequency relation-object pairs from 104 relation-object pairs overlapping between OIV6 validation/testing sets and OOD datasets. We remove all the data samples from this expansion set. For example, if ``holding umbrella'' exists in this set, and one training image has ``person holding umbrella'' with a bounding box for the person and the umbrella, we remove the text and boxes from this training image. Then, we add back OOD data samples with subjects, subjects bounding boxes and relation-object pairs for this expansion set. We conduct experiments on this set to show even without grounded data for these relation-object pairs, we can still get objects in these pairs grounded by adding ungrounded data.

%\textbf{Discussion of Results: } 
Table \ref{table:maintable} shows the main results from our experiments using the training and testing splits defined in Section~\ref{sec:benchmark}. The base model is trained on the {\em Base Training Split}. For our Text-augmented model results, we train another model using the {\em Text-Augmented Training Split}. We report our results on {\em Test A}, {\em Test B}, and {\em Full Test} with R@1 and R@3 for both predicted relation-object pairs and object-location coordinates. %\footnote{R@K with larger K values are reported in supplementary}
%R@K with larger K values are reported in supplementary. However, since over 95\% of all samples have fewer than 3 relation-object pairs per query, so we think R@3 is a sensible level to report without seeing diminishing returns at larger values for K. 
The results in Table~\ref{table:maintable} show considerable improvements to the overall prediction of relation-object pairs and object-locations using our text-augmented model across all tests. 
% Let us dissect the results to see various aspects of this model.
% In Table~\ref{table:maintable}, we follow the same setting: for ``Before'' experiments, we remove 50\% samples belonging to the under-seen set and we remove all the samples belonging to the unseen set during training from grounded data; for ``After'' experiments, we add OOD datasets with ungrounded relation-object pairs for both under-seen and unseen sets. We report our results on under-seen, unseen and all sets with R@1 and R@3 for both predicted text and predicted boxes. 
% For each portion of data samples belonging to enhancement set during training, we find adding back OOD data without bounding boxes annotations for objects can consistently improve the performances for both relation-object generation and grounding. 
%For the relation-object pairs in {\em Rel-Obj Set A}, results for both relation-object generation and bounding box generation are improved, 
These results demonstrate the usefulness of adding text-augmented samples even if they do not have object locations. Suppose that our training set does not have many training samples for {\em cutting orange}, which causes our model to be weaker for this pair. However, our model has already learned how to localize oranges from relation-object pairs such as {\em eating orange} or {\em holding orange}. Any inability to predict {\em cutting orange} is due solely to not seeing enough examples for this relation-object pair. Hence, by simply adding more image-text data to show images with {\em cutting orange}, even if we do not have fully grounded samples, the model can quickly learn to predict this association. %We see that this effect is persistent across other levels of removal, i.e., when more or fewer data points are available for the relation-object pairs in {\em Rel-Obj Set A}.

Next, our results for the relation-object pairs in {\em Test B} are a lot better than the base model. The base results are understandably quite low since this model has not seen any example for relation-object pairs present in {\em Test B}. However, it is remarkable how our model regains the ability to not only predict these pairs but also ground them by solely relying on text-augmented data for these classes. We reiterate that our model has \textit{never seen} any grounded data for these relation-object pairs. Hence, the results come solely from compositional transfer from other pairs and the weak signal from our text augmentation. 
% Both of these results show remarkable potential for the future since the text data is virtually limitless. This also has remarkable prospects for future data collection. We can grow the variety of data much quicker due to how easy it is to train with mixed supervision. In the future, to expand the number and variety of relation-object pairs, we can simply provide box labels for a small amount of additional data, and by combining with a larger amount of ungrounded image-text data. Furthermore, our scheme removes the burden of exhaustively labeling all relationships in each image, we can just sparsely label them over a much wider variety of images instead. For example, instead of labeling 10 relation-object pairs per image, we can label 1 relation-object pair per image for 10 diverse sets of images instead. \kk{Not sure if this belongs here or whether it should go in Future works section....}
While our results already show a huge improvement via text augmentation, we want to emphasize that the augmentation is solely coming from out-of-domain image sources compared to our test set, which makes it even more attractive. However, text augmentation done from an in-domain source image provides even higher performance. We show an experiment comparing in-domain vs out-of-domain ungrounded sample augmentation in the supplementary materials.

\subsection{Comparisons to existing work.}

We have chosen to use a unique input-output paradigm for relationship detection and grounding, which as explained in Section~\ref{sec:method}, offers many benefits and offers a more convenient setup for collecting more data in the future. This setup also allows easy use of image-text data for almost limitless expansion. However, our setup also makes it relatively difficult to put our contributions in the context of existing work. %Does this generation-based method compare favorably with existing methods? We attempt to answer that question here. 
%However, we note that our goal is not to produce the `best' algorithm when trained on a small set of fully-grounded data but rather to 
Here we show that our method is actually comparable to scene graph generation methods and other related methods when using the same data sources.

\vspace{0.05in}
\noindent
\textbf{Comparisons to Scene Graph Generation.} To compare with scene graph generation (SGG) methods, we keep the same 104 relation-object pairs from our {\em Rel-Obj Set A} and {\em Rel-Obj Set B} (since we already have parsed results for text-based augmentation for these). Then, we obtain predictions from both our method and existing work on this set and measure R@K for both relation-object pairs and object-locations. % similar to Section~\ref{sec:exp}.
% \kk{SECTION XXX}. 
However, there are some key differences between our model compared with existing SGG methods. Hence, we have made a few changes in the evaluation to allow for a fair comparison of our model. There are two main concerns:
%\begin{enumerate}%[nolistsep, leftmargin=*]
(1) {\em Our method gets the subject box as input whereas SGG methods do not:} To equalize for this, we only consider predictions from samples where SGG methods correctly predict the subject box. So, for these samples, both methods have the same level of information;
(2) {\em Our method is trained and evaluated on OIv6, whereas SGG methods are only trained on VG:} For this experiment, we trained a model variant on a dataset that excludes all training samples from OIv6 and we only use grounded data from VG + ungrounded expansion from COCO+CC. This reduces the total possible relation-object pairs from 104 to just 28; This will be our test bed for this experiment.
%\end{enumerate}

The results of this experiment are shown in Table~\ref{table:compare_sgg} where we show considerably better results in almost all settings compared to the popular Neural Motif~\cite{zellers2018neural} model for scene graph prediction as well as the recent detector-free model RelTR~\cite{cong2022reltr}. %, except for the box prediction accuracy at IOU of 0.5 for Recall@3. 
%Since our method has generates boxes from scratch, the IOU can be expected to be lower for our method compared to existing SGG models whose pipeline uses many techniques, such as region proposals, to produce higher overlap with object boundaries. Furthermore, our model was not optimized to generate tight bounding boxes %(See Sec. \ref{sec:future} for a brief discussion on this)
%, but rather on whether we can correctly expand the capabilities by adding limited supervision from image-text data. Hence, to fully put our contributions in a better context, 
Since our method is not trained to predict tight bounding boxes using box proposal networks or additional box refinement objectives, we report numbers at varying levels of IOU instead of just a default value of 0.5, showing that a potential combination of our method with box refinement techniques could potentially lead to more improvements.

We observe that our relation-object pair prediction is almost twice as good compared to both of these methods, and at a lower IOU threshold, our method far outperforms box prediction as well. This shows that our method has learned to use text supervision effectively for relation grounding (even if at a coarser grounding threshold, such as IOU $<$ 0.5). While there is some room for improvement in our model in terms of refining the box predictions to better adhere to object boundaries,  
%(See Sec \ref{sec:future})
we believe that these results already show comparable, if not better, results for our model even when compared fairly to traditional scene graph methods.  In addition, our decoder-based unification comes with many other benefits as described earlier. %However, by observing the box recalls with lower IOU, we can see that our method produces boxes that are \textit{close} to the ground truth, which might signify that a lot of improvements can be obtained by just improving the ability of the boxes to snap to object boundaries.

% kk{Please complete methods names here together with why we chose this setup}.

% \kk{I think we should have two different tiny tables for SGG and TAP separately.}

\begin{table}[t!]
\centering
\renewcommand{\arraystretch}{1.2}
\begin{tabular}{l c c c c} 
\toprule
  
 \textbf{Methods} &  \multicolumn{2}{c}{\textbf{Rel-Object}} & \multicolumn{2}{c}{\textbf{Object-Loc}} \\ \cmidrule(l){2-3} \cmidrule(lr){4-5} 
   & R@1 & R@3 & R@1 & R@3 \\
  \midrule
%  TAP+GLIP & 63.55 & xxxx  \\ 
%  TAP + GLIP & 16.31 & 29.18 & 16.03 & 28.69  \\
 TAP~\cite{pham2022improving} + GLIP & 16.31 & 29.18 & 11.64 & 21.31  \\ 
%  Ours+GLIP & - & xxxx  \\ 
 \OCRDNet (Ours) & 63.16&91.53 & 49.50&71.46 \\
 
 \bottomrule
 \end{tabular}
 \vspace{-0.08in}
 \caption{We compare our results against a specialized attribute prediction model combined with an open-vocabulary object detection model (GLIP~\cite{li2022grounded}) on 100 relation-object pairs in from the Full Test set that are in the vocabulary for TAP.}
 \label{table:compare_tap}
 \vspace{-0.2in}
 \end{table}

\begin{figure*}[t]
\vspace{-0.3in}
\begin{center}
% \vspace{-5in}
\includegraphics[width=0.9\textwidth]{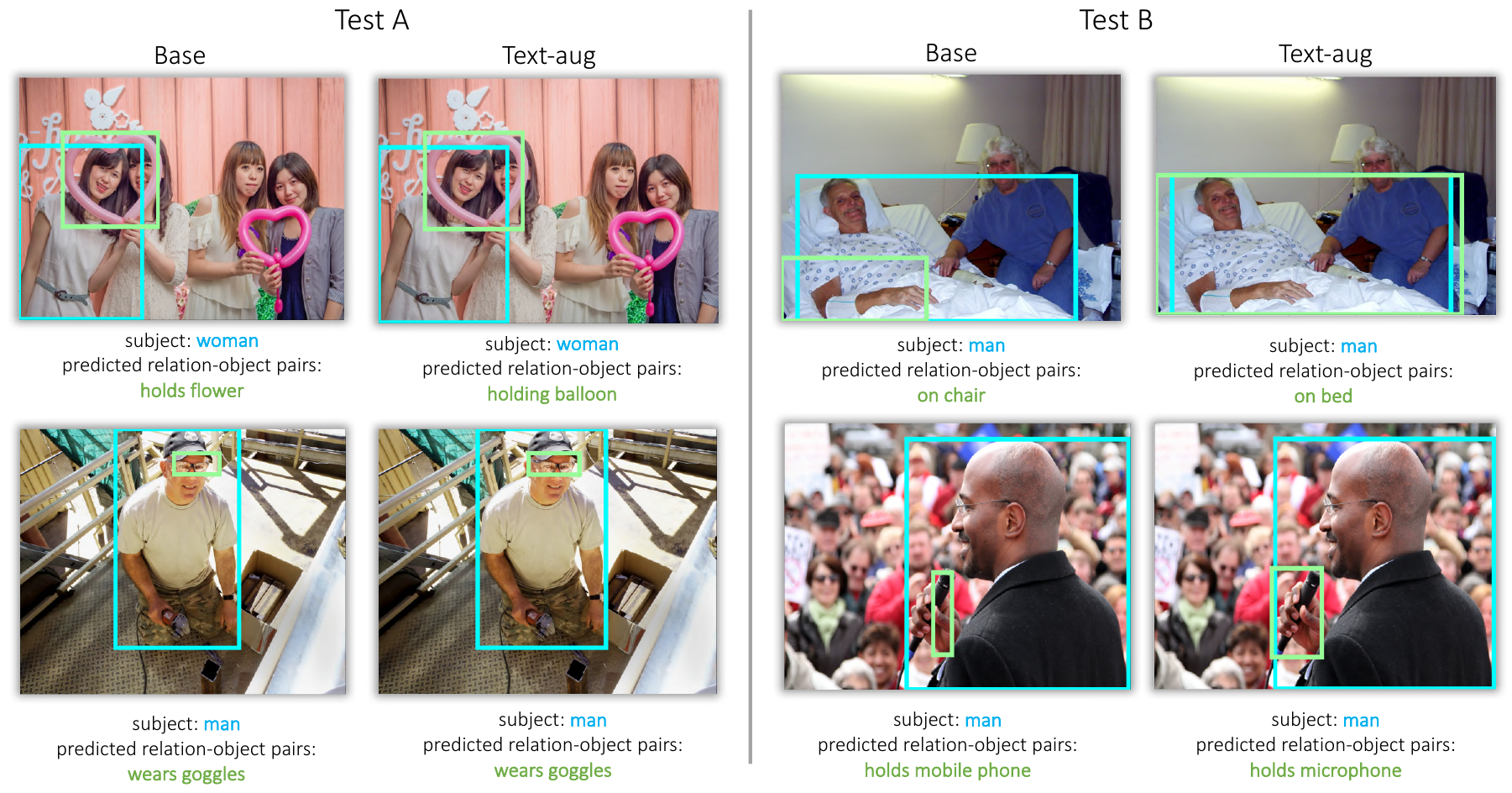}
\vspace{-0.1in}

\caption{Qualitative results from the models trained with and without text augmentation from COCO and CC. ``Base'' indicates results generated by the model trained with 50\% training samples for the relation-object pairs in {\em Rel-Obj Set A} and no samples for the relation-object pairs in {\em Rel-Obj Set B}. ``Text-aug.'' indicates results generated by the model trained with Text-Augmented Training Split.}
\label{fig:qual_improve}
\end{center}
\vspace{-0.25in}
\end{figure*}

\vspace{0.05in}
\noindent
\textbf{Other Comparisons.} 
Another closely related work is the recently proposed TAP~\cite{pham2022improving} model for attribute prediction. This model is trained to predict related attributes for a given query of object names and boxes which is similar to our setup. This work also considers relation-object pairs as attributes of an object, so it is possible to obtain relation-object pair predictions from this model. By combining TAP with GLIP~\cite{li2022grounded}, we can obtain both text and box prediction. If GLIP generates multiple bounding boxes for one object, we keep the box closest to the ground truth subject bounding box. We find that TAP can predict 100 relation-object pairs in the Full Test set, and we report results on these 100 relation-object pairs in Table~\ref{table:compare_tap}. We find that in both comparison settings, our work can far outperform this potential solution in both text and box predictions. The TAP model was generously provided by the original authors in order to conduct this experiment.

\subsection{Ablation Experiments}

\vspace{0.05in}
\noindent\textbf{Effect of adding more image-text data.}
\label{subsec:ats}
Our earlier results have already shown that adding text data can help improve prediction and grounding for both previously seen and unseen relation-object pairs. Here we want to explore how adding more data over time might affect this. To do this, we set our evaluation split to only contain the samples from Test A + Test B. Then, our base training split is constructed so that all the training samples of the 104 relation-object pairs belonging to {\em Rel-Obj Set A} and {\em Rel-Obj Set B} are removed. We refer to this ablation training set as ATS. Then, we incrementally add image-text augmentation with COCO and CC15M. The results are presented in Table~\ref{table:data_ablation}, which clearly shows that the inclusion of more data has additive effects. This suggests that our model can very effectively scale to include new relation-object pairs that do not have any grounded data during training, and we can anticipate it to continue getting better with more samples from vast collections of image-text pairs on the internet. For example, resources such as LAION-2B and LAION-400M~\cite{schuhmann2021laion}, COYO-700M~\cite{kakaobrain2022coyo-700m}, and RedCaps~\cite{desai2021redcaps} can potentially be used which can add hundreds of millions of additional text-augmentation.

% \vspace{-0.1in}
Additional ablation studies are included in the supplementary material such as studying the effect of removing more samples from the relation-object pairs from {\em Rel-Obj Set A}. Currently we remove 50\% of the samples but we show that our results hold when removing less 25\% or more samples 75\%. Finally, we show in Figure~\ref{fig:qual_improve} some qualitative results for relation-object pairs and object-locations predicted by our models with and without text augmentation. 
% Additional qualitative results are included in the supplementary material. 
% We plan to release all training and test splits as well as code upon paper acceptance.

\begin{table}[t!]
\centering
\setlength\tabcolsep{4.6pt}
\renewcommand{\arraystretch}{1.2}
\begin{tabular}{l c c c c} 
\toprule
   \textbf{Data} & \multicolumn{2}{c}{\textbf{Rel-Object}} & \multicolumn{2}{c}{\textbf{Object-Loc}} \\ \cmidrule(l){2-3} \cmidrule(lr){4-5} 
   & R@1 & R@3 & R@1 & R@3 \\
  \midrule
 ATS & 0.64 & 3.87 & 0.62&2.82 \\ 
 ATS + COCO & 8.74&20.24 & 6.74&13.95 \\
 ATS + COCO + CC & 20.96&37.59 & 12.20&20.74 \\
%  \hline
%  ATS & 0.64/3.87/7.13 & 0.62/2.82/4.96 \\ 
%  ATS+COCO & 8.74/20.24/28.01 & 6.74/13.95/18.30 \\
%  ATS+COCO+CC & 20.96/37.59/44.34 & 12.20/20.74/24.15 \\

%  Data & R@1,3,5 Text & R@1,3,5 Box \\
 
%  \hline
%  ATS & 0.64/3.87/7.13 & 0.62/2.82/4.96 \\ 
%  ATS+COCO & 20.93 & 14.31 \\
%  ATS+COCO+CC15M & 20.96/37.59/44.34 & 12.20/20.74/24.15 \\

% beam 5, seq 3: R@3 42.29, 23.30
%  \hline 
\bottomrule
\end{tabular}
\vspace{-0.05in}
\caption{Effect of additional text augmentation from image captions, starting from a base training on the Ablation Training Set (ATS) (See~Sec.~\ref{subsec:ats}). Considerable gains are obtained by text-augmented data without any additional box information.}
\label{table:data_ablation}
\vspace{-0.2in}
\end{table}

% \zy{we don't have such results yet, I'm not sure if we should add this conclusion}
% We conduct an ablation study for our model in two aspects: 1) Different levels of data removal to test improvement of under-seen samples during training, and 2) The effect of including more image-text data on relationship prediction. The results for these experiments are shown in Tables XXX and XXX. 

% to answer two main questions: 1) When we change the number of samples of strongly grounded training samples seen during training, how does it affect the ability of our image-text augmentation to 

% two aspects: first, we want to explore whether image-text augmentation can help relationship prediction and grounding when different portions of samples hav

% different portions of removal samples for under-seen relationships. Second, w

\vspace{-0.05in}
\section{Conclusion}
\label{sec:future}

The \OCRD task combined with generation-based relation detection offers several advantages. To expand the number and variety of relation-object pairs, we can simply provide sparse box annotations for a small amount of additional data, and combine them with a larger amount of ungrounded image-text data to obtain the same benefits as annotating a lot of expensive box annotations. 
% Furthermore, our scheme removes the burden of exhaustively labeling all relation-object pairs in each image, we can just sparsely label them over a much wider variety of images instead. For example, instead of labeling 10 relation-object pairs per image, we can label 1 relation-object pair per image for 10 diverse sets of images instead, which can scale more effectively.
%Beyond the big picture, there is also room for improvements to our model, the biggest of which is our box prediction accuracy, which lags behind our impressive numbers for predicting the right relation-object pairs. However, by observing the box recalls with lower IOU, we can see that our method produces boxes that are \textit{close} to the ground truth, which might signify that a lot of improvements can be obtained by just improving the ability of the boxes to snap to object boundaries. One solution is to include a sense of distance in addition to the simple token-based decoding for box corners. Next, we believe the granularity of the tokens used for boxes is also \textit{too fine}, which penalizes the model for even small deviations. While both of these things are fixable, they do require non-trivial innovations and re-training of the backbone models. 
% However, our main goal for this paper was to introduce \OCRD setup and show the benefits of using additional image-text pairs on the task of relation prediction. 
In summary, we proposed subject-conditional relation detection (\OCRD) which is an alternative way to obtain a relation graph given a query object. We also propose a new transformer-based model which allows us to effectively tackle this task. \OCRD, combined with our method allows us to easily utilize image-text paired data to further enhance relation prediction, even when there are no annotations present in the base training set for these relationships. 

\paragraph{Acknowledgments} This work was partially supported by gifts from Adobe Research and NSF Awards IIS-2201710 and IIS-2221943.

%%%%%%%%% REFERENCES
{\small
\bibliographystyle{ieee_fullname}
\bibliography{egbib}
}

\section{Supplementary}

In this section, we provide more details on R@K with larger K (K = 5, K =10) and the influence of in-domain text augmentation. We also show more qualitative examples of our proposed model with and without text augmentation from image captions. 

\subsection{Larger K values}
\begin{table}[h]
\centering

\small
\renewcommand{\arraystretch}{1.2}
\setlength\tabcolsep{3.2pt}

% \small
% \renewcommand{\arraystretch}{1.3}
% \setlength\tabcolsep{5pt}
\begin{tabular}{llcccc}
\toprule
\multirow{2}{*}{\textbf{Remov. \%}} &     \multirow{2}{*}{\textbf{Recall}}     & \multicolumn{2}{c}{\textbf{Test A}} 
& \multicolumn{2}{c}{\textbf{Full Test}} \\ \cmidrule(r){3-4} \cmidrule(lr){5-6} 
                        &          & Base                    & Text-aug.                    & Base                 & Text-aug.                 \\
\midrule
% \multirow{4}{*}{25\%} & Rel-Obj R@5 &       86.58                   &             87.51          &          71.71             &    81.24                   \\
%                         & Rel-Obj R@10 &       88.68                  &            89.37         &          73.87            &    84.34   \\

%                         \cline{2-6} 
%                         & Obj-Loc R@5  &    60.50                    &               61.27         &         52.61               &        59.60               \\ 
%                         & Obj-Loc R@10  &    62.36                   &               63.16        &         54.57               &        61.97
%                         \\
                        
%                         \midrule
\multirow{4}{*}{50\%}   & Rel-Obj R@5 &        84.54                   &         86.17                 &            71.35            &        78.18  
                        \\ 
                        & Rel-Obj R@10 &        87.35                  &         88.57               &            73.95            &        81.38 
                        \\
                        
                        \cline{2-6} 
                        & Obj-Loc R@5  &          58.58                 &          59.90                &            52.18            &   57.37                    \\ 
                        & Obj-Loc R@10  &          60.36                &          61.71                &            54.38           &   59.87
                        \\
                        
%                         \hline
% \multirow{4}{*}{75\%}   & Rel-Obj R@5 &       79.99                    &        83.20                 &         70.93               &  78.56  
%                         \\ 
%                         & Rel-Obj R@10 &       83.20                   &        86.53                 &         73.85               &  81.95 
%                         \\
                        
%                         \cline{2-6} 
%                         & Obj-Loc R@5  &       53.21                   &        55.89                  &          51.35              &  56.87                    \\
%                         & Obj-Loc R@10  &       55.48                   &        58.48                &          53.59             &  59.29  
%                         \\
                        \bottomrule
                           
\end{tabular}
% \vspace{-0.1in}
% \vspace{0.15in}

\caption{Results with different $K$ values for Test A and Full Test. “Base” indicates removing different portions of training samples for the relation-object pairs in {\em Rel-Obj Set A} and removing all training samples for {\em Rel-Obj Set B}. "Text-aug." indicates adding ungrounded samples for both unseen and under-sampled relation-object pairs.}
\label{table:supp_K}
\vspace{-0.10in}
\end{table}

\begin{table}[t]
\centering
\small
\renewcommand{\arraystretch}{1.2}
\setlength\tabcolsep{3.2pt}
\begin{tabular}{llcccc}
\toprule
\multirow{2}{*}{\textbf{Remov. \%}} &     \multirow{2}{*}{\textbf{Recall}}     & \multicolumn{2}{c}{\textbf{Test A}} 
& \multicolumn{2}{c}{\textbf{Full Test}} \\ \cmidrule(r){3-4} \cmidrule(lr){5-6} 
                        &          & Base                    & Text-aug.                    & Base                 & Text-aug.                 \\
\midrule
\multirow{4}{*}{25\%}   & Rel-Obj R@1 &     46.61                     &          49.28               &        50.45                &          54.14             \\
                        & Rel-Obj R@3 &       84.75                   &            85.34         &          69.80            &    75.83   \\
                        % & Rel-Obj R@5 &       86.58                   &             87.51          &          71.71             &    81.24                   \\
                        \cmidrule{2-6} 
                        & Obj-Loc R@1  &      33.22                     &      35.06                    &             37.18           &  39.80   
                        \\
                        & Obj-Loc R@3  &    59.51                   &               59.92        &         51.13               &        55.64 
                        \\
                        % & Obj-Loc R@5  &    60.50                    &               61.27         &         52.61               &        59.60               \\ 
                        \midrule
\multirow{4}{*}{50\%}   & Rel-Obj R@1 &        39.94                   &         43.61                 &        49.70                &                53.81      \\
                        & Rel-Obj R@3 &        82.19                  &         83.91               &            69.22            &        75.51 
                        \\
                        % & Rel-Obj R@5 &        84.10                   &         86.43                 &            71.28            &        80.47  
                        % \\ 
                        \cmidrule{2-6} 
                        & Obj-Loc R@1  &        29.08                   &     31.75                   &          36.54              &          39.36   
                        \\
                        & Obj-Loc R@3  &          57.06                &          58.58               &            50.37           &   55.38
                        \\
                        % & Obj-Loc R@5  &          58.17                 &          60.39                &            52.04            &   58.73                    \\ 
                        \midrule
\multirow{4}{*}{75\%}   & Rel-Obj R@1 &     32.37                      &                37.90          &       49.04                 &               53.09        \\
                        & Rel-Obj R@3 &       77.56                   &        80.43                  &         68.80               &  75.18 
                        \\
                        % & Rel-Obj R@5 &       79.99                    &        83.20                  &         70.93               &  78.56  
                        % \\ 
                        \cmidrule{2-6} 
                        & Obj-Loc R@1  &     25.05                      &        27.01                  &                36.26        &  38.84                     \\
                        & Obj-Loc R@3  &       52.23                   &        54.32                 &          49.84             &  54.55   
                        \\
                        \bottomrule
                        % & Obj-Loc R@5  &       53.21                   &        55.89                  &          51.35              &  56.87   
\end{tabular}
% \vspace{-0.05in}

\vspace{0.08in}
\caption{Results for Test A and Full Test. “Base” indicates removing different portions of training samples for the relation-object pairs in {\em Rel-Obj Set A} and removing all training samples for {\em Rel-Obj Set B}. "Text-aug." indicates adding ungrounded samples for both unseen and under-sampled relation-object pairs.}
\label{table:ablation-percent}
\vspace{-0.10in}
\end{table}

In Table~\ref{table:supp_K}, we report R@5 and R@10 results for Test A and Full Test. Increasing K improves both relation-object prediction and object-location coordinates prediction, but such benefits start to become marginal when K is larger. For example, from Table~\ref{table:maintable}, the Rel-Obj recall and Obj-Loc recall increase to 75.51 (Rel-Obj R@3) and 55.38 (Obj-Loc R@3) from 53.48 (Rel-Obj R@1) and 39.36 (Obj-Loc R@1) when K value increases from 1 to 3. By considering two more predictions for all the relation-object pairs, Rel-Obj recall increases 22.03, and Obj-Loc recall increases 16.02. However, Obj-Loc @10 and Rel-Obj @10 can reach 81.38 and 59.87 from 75.51 (Rel-Obj R@3) and 55.38 (Obj-Loc R@3) by considering seven more predictions, showing a 5.87 improvement in Rel-Obj recall and 4.49 improvement in Obj-Loc recall.

\subsection{What is the effect of the amount of grounded data seen during base training?}

In Table~\ref{table:ablation-percent}, we show the numbers of different amounts of data `removed' from triplets belonging to {\em Test A}. This setting simulates the presence of relation-object pairs that might not have many training samples with strong grounding annotations. As seen in Table~\ref{table:ablation-percent}, our method continues to add additional value at all levels of removal. With text augmentation, our method compares very favorably with having a lot more grounded data in the base training. For example, after text augmentation, our model with 50\% data removed has a slightly better relation-object prediction and almost the same object-location coordinates prediction as the `Base' model trained with only 25\% data removed ($\sim$24k additional fully grounded training samples).
% \kk{ZIyan fill this XXX}). 
This further suggests a strong possibility of rapid expansion of the total number of unique relation-object pairs by only annotating a small amount of additional grounded data, and using a large amount of image-text data to obtain the same benefit as annotating a lot of expensive grounding data.

\subsection{In-domain Text Augmentation}
\begin{table}[h]
\centering
\setlength\tabcolsep{4pt}
\begin{tabular}{l c c c c} 
\toprule
   \textbf{Data} & \multicolumn{2}{c}{\textbf{Rel-Object}} & \multicolumn{2}{c}{\textbf{Object-Loc}} \\ \cmidrule(l){2-3} \cmidrule(lr){4-5} 
   & R@1 & R@3 & R@1 & R@3 \\
  \midrule
 ATS + OutDomain & 20.96 & 37.59 & 12.20 & 20.74 \\
 ATS + InDomain  & 48.24 & 86.51 & 23.93 & 43.73 \\

\bottomrule
\end{tabular}
% \vspace{0.15in}
\caption{Effect of additional text augmentation from in-domain data (OIv6+VG+Flickr) and out-of-domain data (COCO+CC), starting from a base training on the Ablation Training Set (ATS) (See~Sec.~\ref{subsec:ats}). Considerable gains are obtained by text-augmented data without any additional box information. Results are reported for 104 relation-object pairs.}
\label{table:supp_indomain}
\vspace{-0.15in}
\end{table}

\begin{figure*}[h!]
\begin{center}
% \vspace{-1in}
\includegraphics[width=0.93\textwidth]{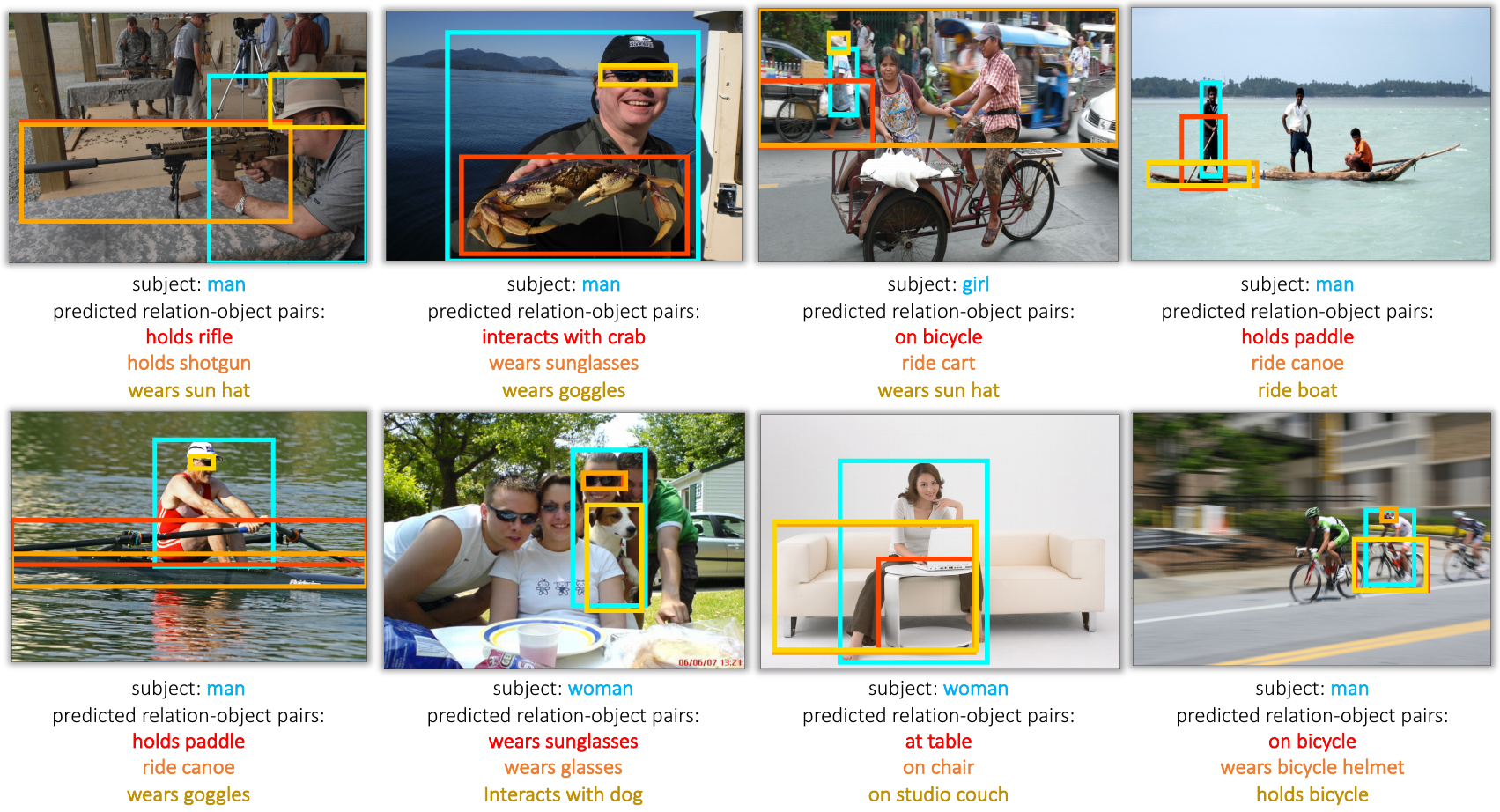}
% \vspace{0.05in}

\caption{Qualitative results from the model trained with fully grounded data. Subjects and predicted relation-object pairs are shown under images, and bounding boxes correspond to subjects and objects with the same colors. During inference time, all subjects and bounding boxes for subjects are provided as inputs, and we show three predicted relation-object pairs with highest scores.}
\label{fig:qual_fully_supp}
\end{center}
% \vspace{-1.0in}
\end{figure*}

In Table~\ref{table:supp_indomain}, we report results for 104 relation-object pairs from Test A and Test B. For both experiments, we remove all the samples for these 104 pairs from VG, Flickr30k and OIv6. Then, we add ungrounded samples for these 104 relation-object pairs from VG+Flickr30k+OIv6 (in-domain data) and from COCO+CC (out-of-domain data) separately. In-domain data contains clean subjects and relation-object pairs for images, and shows higher improvement than out-of-domain data for both text and box predictions as expected.

\subsection{Qualitative Examples}
First, we show predicted relation-object pairs and bounding boxes for objects using fully grounded data. In this case, the model is trained by Visual Genome, Flickr30k and OIv6 with all the text and bounding boxes. In Figure~\ref{fig:qual_fully_supp}, our model can successfully predict correct relation-object pairs given a specific subject and the bounding box for the subject. For example, in the top right figure, by given the text and location of the leftmost man, the model can successfully predict ``holds paddle'' and ``ride canoe'' with corresponding bounding boxes for ``paddle'' and ``canoe'' for the given man, even there is another paddle in the same figure. These examples also show the future direction to improve the grounding performance. For example, in the second row, the third image shows the predicted boxes for ``chair'' and ``studio couch'', but these boxes are not accurate enough because of the incorrect prediction for the bottom-right x-coordinate. Since our model localize objects by generating box coordinates directly, the generated boxes can be improved by other post-processing steps to refine box predictions.

\begin{figure*}[h!]
\begin{center}
% \vspace{-5in}
\includegraphics[width=0.95\textwidth]{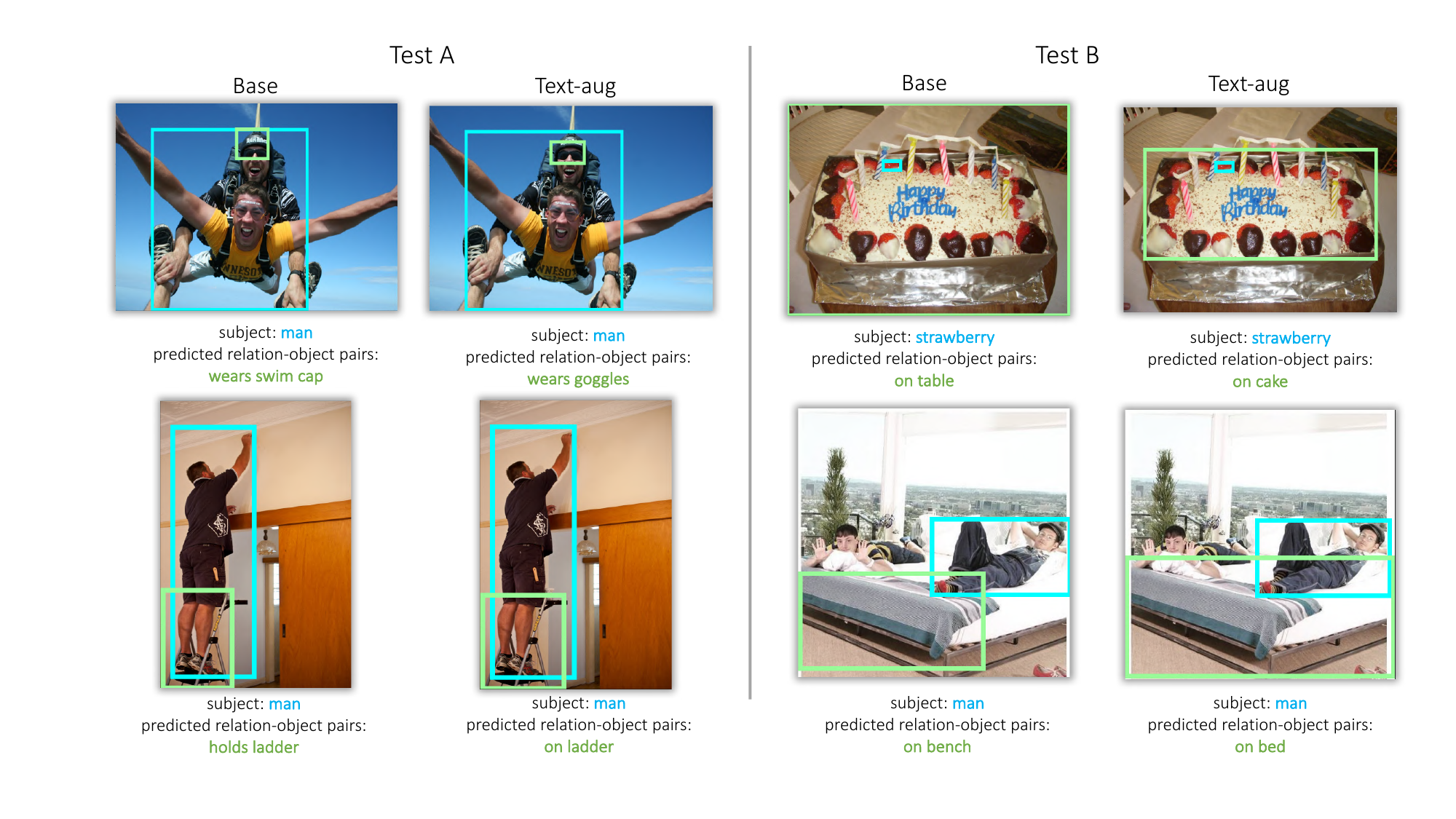}
% \vspace{-0.05in}

\caption{Qualitative results from the models trained with and without text augmentation from COCO and CC. ``Base'' indicates results generated by the model trained with 50\% training samples for the relation-object pairs in {\em Rel-Obj Set A} and no samples for the relation-object pairs in {\em Rel-Obj Set B}. ``Text-aug'' indicates results generated by the model trained with additional ungrounded data for both relation-object pairs in {\em Rel-Obj Set A} and {\em Rel-Obj Set B} from COCO and CC.}
\label{fig:qual_improve_supp}
\end{center}
\vspace{-0.1in}
\end{figure*}

In Figure~\ref{fig:qual_improve_supp}, results for ``Base'' are generated by the model trained with VG+Flickr30k+OIv6 by removing 50\% of training samples for relation-object pairs in {\em Rel-Obj Set A} and removing all training samples for pairs in {\em Rel-Obj Set B}. The model generating results for ``Text-aug'' is trained with the same amount of grounded data and additional ungrounded data from COCO+CC for the relation-object pairs in {\em Rel-Obj Set A} and {\em Rel-Obj Set B}. For both sets, additional text augmentation helps the text prediction and grounding. For example, in the top left example, without additional text augmentation, the relation-object pair ``wears goggles'' cannot be predicted correctly. Text augmentation introduces more images to help the model to recognize relation-object pairs better. 

% the relation-object pairs in {\em Rel-Obj Set A} and no samples for the relation-object pairs in {\em Rel-Obj Set B}

\end{document}